\documentclass{article}

\usepackage{arxiv}

\usepackage[utf8]{inputenc} 
\usepackage[T1]{fontenc}    
\usepackage{hyperref}       
\usepackage{url}            
\usepackage{booktabs}       
\usepackage{amsfonts}       
\usepackage{nicefrac}       
\usepackage{microtype}      
\usepackage{lipsum}
\usepackage{amsmath,graphicx,subfig}

\title{Dropout Regularization in\\ Hierarchical Mixture of Experts}

\author{
  Ozan \.Irsoy\\
    Bloomberg L.P.\\
       731 Lexington Ave.\\
       New York, NY 10022, USA\\
  \texttt{oirsoy@bloomberg.net} \\
   \And
   Ethem Alpayd{\i}n\\
  Department of Computer Engineering,\\
       Bo{\u g}azi{\c c}i University\\
       TR-34342 Istanbul, Turkey \\
  \texttt{alpaydin@boun.edu.tr} \\
}

\begin{document}
\maketitle

\begin{abstract}
Dropout is a very effective method in preventing overfitting and has become the go-to regularizer for multi-layer neural networks in recent years. Hierarchical mixture of experts is a hierarchically gated model that defines a soft decision tree where leaves correspond to experts and decision nodes correspond to gating models that softly choose between its children, and as such, the model defines a soft hierarchical partitioning of the input space. In this work, we propose a variant of dropout for hierarchical mixture of experts that is faithful to the tree hierarchy defined by the model, as opposed to having a flat, unitwise independent application of dropout as one has with multi-layer perceptrons. We show that on a synthetic regression data and on MNIST and CIFAR-10 datasets, our proposed dropout mechanism prevents overfitting on trees with many levels improving generalization and providing smoother fits.
\end{abstract}


\section{Introduction}

In machine learning, we care about the generalization performance of our prediction function over unseen data. Regularization is one of the fundamental methods to combat overfitting on the training data, by preferring smaller, simpler, or smoother models. Traditional regularizers work by pushing model parameters to smaller values (e.g., L1/L2 regularizers), or increasing the model sparsity.

Deep neural networks composed of many layers with many units can overfit and a number of regularization methods have been proposed. One such method is the dropout in which each unit is randomly dropped (along with its connections) from the network with a certain probability~\citep{JMLR:v15:srivastava14a}. Intuitively, this reduces the co-adaptation of units by forcing them to not rely on one another too much. It is also equivalent to sampling from an exponential number of networks at training time. Dropout method has spurred research and several variants have been proposed, such as drop-connect where each connection can be separately dropped~\citep{wan2013regularization}, fast-dropout which performs a Gaussian approximation to the implied objective~\citep{wang2013fast}, and word-dropout in natural language processing where each word in a given sentence is dropped out (or replaced with the unknown token)~\citep{iyyer2015deep}. Later work grounded the approach in a stronger theoretical basis~\citep{gal2016dropout,gal2016theoretically}.

Hierarchical mixture of experts (HMoE) is a meta-model that combines several models using a gating function that is defined with respect to a tree hierarchy~\citep{jordan1994hierarchical}. When the gating function and the individual experts are differentiable with respect to their parameters, the whole tree can be trained using stochastic gradient-descent (SGD), just like we train the weights of a neural network given its structure. Soft decision trees~\citep{irsoy2012soft} extend HMoEs by learning the tree structure as well, using a greedy incremental method similar to how classical decision trees are trained. In a more recent extension titled Budding Trees~\citep{irsoy2014budding}, both the parameters as well as the tree structure is learned using SGD. 

When the hierarchy has many levels with the number of gating models and experts increasing exponentially with depth, HMoE is also prone to overfitting and in this work, we propose a dropout mechanism that is suitable to the tree structure. Our experiments show that our proposed dropout method works as a regularizer and improves generalization. In this work we focus on applying the proposed dropout method to HMoEs for simplicity and ease of exposition, however the method is readily applicable to soft decision trees or budding trees as well.

\section{Preliminaries}

\subsection{Hierarchical Mixture of Experts}

{\bf Mixture of experts} architecture consists of multiple experts and a gating model. The gating model is basically a classifier that divides up the input space among the experts and each expert is responsible for generating the correct output in its domain of expertise (as defined by the gating model)~\citep{jacobs1991adaptive}. More formally, let $f_i(x)$ be the output of expert $i$ for input $x$ and $\alpha(x)$ denote the gating function. Then, the overall response is calculated as:
\begin{align}
y(x) = \sum_{i=1}^K \alpha_i(x) f_i(x)
\end{align}
where $\sum_i \alpha_i(x) = 1$ and there are $K$ experts. A simple gating function would divide the subspace linearly:
\begin{align}
\alpha(x) = \text{softmax}(w^T x)
\end{align}

\noindent which reduces to the sigmoid when $K=2$. Another interpretation of this model is that $\alpha_i(x)$ models the probability of $x$ falling into the $i$th subspace (where subspaces are linearly divided in the particular gating function above), and $y(x)$ computes the expected response function with respect to this probability distribution.

If we replace each expert by a mixture of experts, we get the {\bf hierarchical mixture of experts}~\citep{jordan1994hierarchical}. The recursion defines a tree hierarchy with gating models working as internal decision nodes splitting the input space recursively and experts at the lowest level correspond to the leaves. This defines a {\em soft decision tree\/} because the splitting is not hard: Unlike a hard decision tree where we take one path from the root to one of the leaves, we traverse all paths to all the leaves and we take a sum weighted by the gating values on each path. More formally,
we have:
\begin{align}
    y_m(x) &= \begin{cases}
    \alpha_m(x) y_{ml}(x) + (1-\alpha_m(x)) y_{mr}(x) \hfill \hspace{10pt} &\text{ if $m$ is internal}\\
                   f_m(x) \hfill \hspace{10pt} &\text{ if $m$ is leaf}
    \end{cases}\\
    \alpha_m(x) &= \text{sigmoid}(w_m^T x)
\end{align}

Here we assume branching into two and we have a binary tree. For node $m$, $ml$ and $mr$ denote its left and right children respectively. $\alpha_m()$ denotes the local gating function for node $m$.

In the simplest case, $f_m(x)$ is a constant value, i.e., $f_m(x) = c_m$ (which could be a vector value, such as log-odds for each class in a classification task). In this case, hierarchical mixture of experts is similar to a feedforward neural network with one hidden layer, with the difference that there is a functional dependency between each unit determined by the tree structure.

\subsection{Dropout}

Dropout is a regularization method that reduces the  co-adaptation of feature learners in a network. It works by  randomly dropping out units in a neural network, therefore the network has to learn to work in the absence of each of its units. 

Let us limit our scope to a feedforward layer. A single hidden unit $h_i$ would be defined as:
\begin{align}
    h_i = \sigma (w_i^T x)
\end{align}
where $\sigma(\cdot)$ denotes a squashing nonlinearity such as tanh or sigmoid. If we apply dropout to the hidden layer with a probability value of $p$, the equation would be modified as follows (at training time):
\begin{align}
    d_i &\sim \text{Bernoulli}(1-p)\\
    h_i &= d_i \cdot \sigma (w_i^T x)
\end{align}
This shows that with probability $p$, we assign $h_i$ to be zero, effectively dropping that unit out from the network. In recent work, dropout has shown to improve generalization ability across different architectures and different tasks, and has gained widespread use~\citep{JMLR:v15:srivastava14a,dahl2013improving,
krizhevsky2012imagenet,kim2014convolutional,zaremba2014recurrent,irsoy2014opinion}.

\section{A Dropout Method for Hierarchically Gated Models}

In hierarchical mixture of experts, we do dropout on gating units. Again for each node $m$, we sample from a Bernoulli using a dropout-rate hyperparameter. If we choose to drop, this results in dropping out the entire left subtree, i.e. resorting to only the right subtree for the response function. For the internal node $m$:
\begin{align}
d_m &\sim \text{Bernoulli}(1-p)\\
y_m(x) &= \begin{cases}
\alpha_m(x) y_{ml}(x) + (1-\alpha_m(x)) y_{mr}(x) \hfill \hspace{10pt} \text{ if } d_m = 0\\
               y_{mr}(x) \hfill \hspace{10pt} \text{ if } d_m = 1
\end{cases}
\end{align}

At training time, dropout implies (randomly) setting $\alpha_m$ of left to 0, which implies $\alpha_m$ of 1 for the right child.
At test time we do not rescale with $(1-p)$ as in the traditional dropout~\citep{JMLR:v15:srivastava14a}. This is because in both dropped and non-dropped cases, the gating values sum to 1.

This approach is different from the traditional dropout used on perceptrons in which we drop or not a single unit independently of other units. Hierarchical mixture of experts connects each unit to one another in a hierarchical gating structure where each parent controls the gating value assigned to its children, in contrast to the flat structure defined by the perceptron layer. Dropout approach being presented here respects this hierarchy: Dropping out one of the children results in turning off the entire path of responsibility through that child node, resulting in dropping out an entire subtree.

The reason we break the symmetry and drop only the left subtree as opposed to randomly dropping either is as follows: Randomly dropping either subtree would push both subtrees to implement similar functionalities using the same complexity. This would be equivalent to ensembling trees with half the number of nodes. Dropping only the left subtree pushes the right subtree to implement a similar function to the whole tree, while having fewer nodes. This is equivalent to ensembling trees of different complexities.

\section{Experiments}

\subsection{Setting}

We limit the tree structure of hierarchical mixture of experts to a complete binary tree of given depth. The depth of the tree is a hyperparameter and we investigate the impact of dropout for increasing depths. We use the cross-entropy loss  (with softmax nonlinearity at the top) for classification and squared loss for regression. Training is done using minibatched stochastic gradient descent with Adam update rule~\citep{kingma2014adam} on a fixed structure where we update all parameters (in gatings on all levels and leaves) simultaneously. We do not apply any other regularization (such as weight decay or norm-based regularizers such as L1 or L2) in addition to the dropout presented in this work.

The dropout rate values that we investigate are typically small values such as 0.05 or 0.1, as opposed to values used in traditional dropout, such as 0.5~\citep{JMLR:v15:srivastava14a}. This is because in the dropout on a tree, the expected number of retained units decrease exponentially with the depth of the tree. A dropout value of 0.5 would mean that at every level we trim approximately half of the remaining subtree, whereas in a flat feedforward architecture, this value would approximately half of the overall number of units. We could, alternatively, use a dropout rate which increases with the depth of the tree, however scheduling the rate as a function of depth would bring in additional hyperparameter tuning complexity.

\begin{figure}[t]
    \centering
    \subfloat[No dropout]{\includegraphics[width=.33\textwidth]{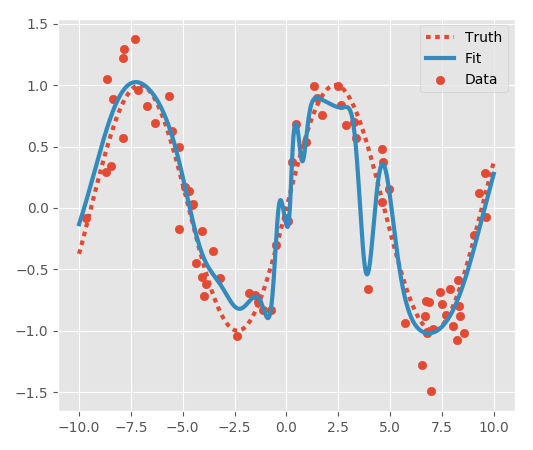}}%
    \subfloat[$p=0.05$]{\includegraphics[width=.33\textwidth]{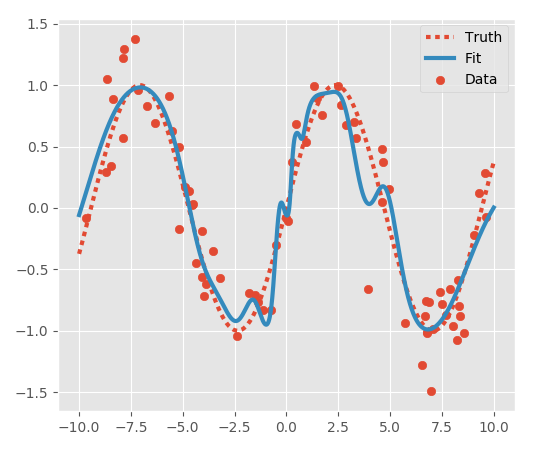}}%
    \subfloat[$p=0.2$]{\includegraphics[width=.33\textwidth]{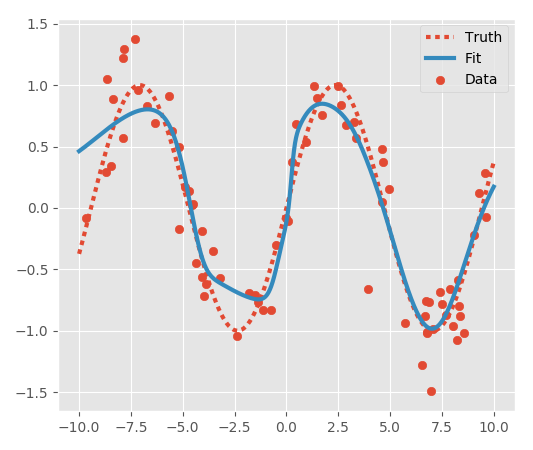}}\\
    \subfloat[$p=0.3$]{\includegraphics[width=.33\textwidth]{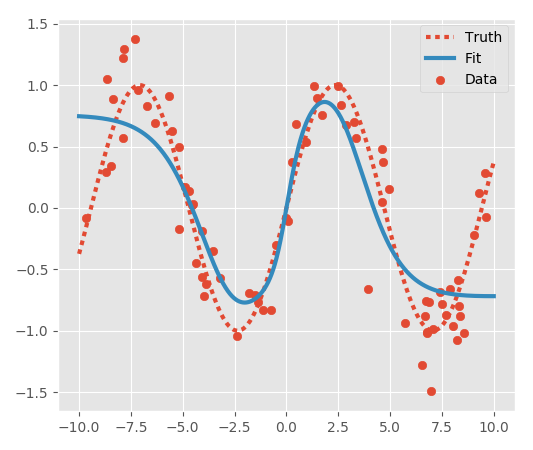}}%
    \subfloat[$p=0.5$]{\includegraphics[width=.33\textwidth]{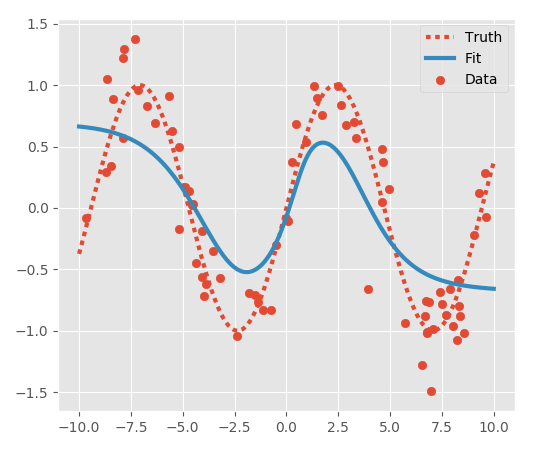}}%
    \subfloat[$p=0.8$]{\includegraphics[width=.33\textwidth]{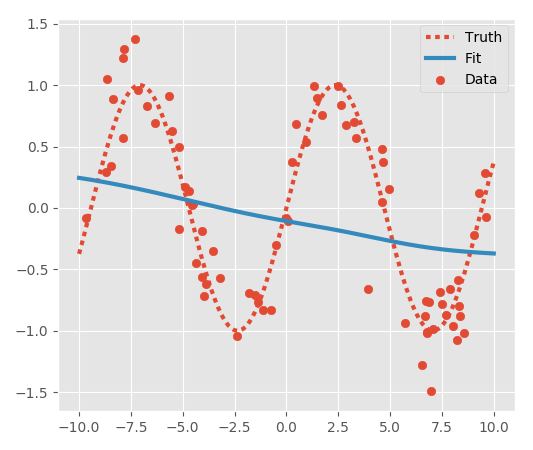}}%
    \caption{Fits using a HMoE model with depth 10 and 1023 internal nodes for various dropout rates. We see decreasing variance, increasing bias and smoothness as we increase the dropout rate.}
    \label{fig:toy}
\end{figure}

\subsection{Results on Toy Data}

We test our approach on a one-dimensional synthetic dataset, which is sampled from a sinusoid function with small Normal noise. The task is posed as a regression problem with single scalar response value, therefore we use the squared loss without any nonlinearity at the output of the model. We use a learning rate of $10^{-3}$ and run a model with 1023 internal nodes (depth 10) for 1000 epochs.

Resulting fits are shown in Figure~\ref{fig:toy} for different values of the dropout rate $p$. We observe that for the case of no dropout as well as for small values of dropout rates, such as 0.05, the model tends to overfit; i.e., it learns the noise and exhibits steep jumps. As we increase the dropout rate, the fit becomes smooth, eventually to the point where we get almost a linear fit with very low variance and high bias. This indicates that our proposed method works effectively as a regularizer with the dropout rate as its hyperparameter.

\subsection{Results on Digit Recognition}

\begin{figure}[t]
    \centering
    \subfloat[$l=4095$, no dropout]{\includegraphics[width=.33\textwidth]{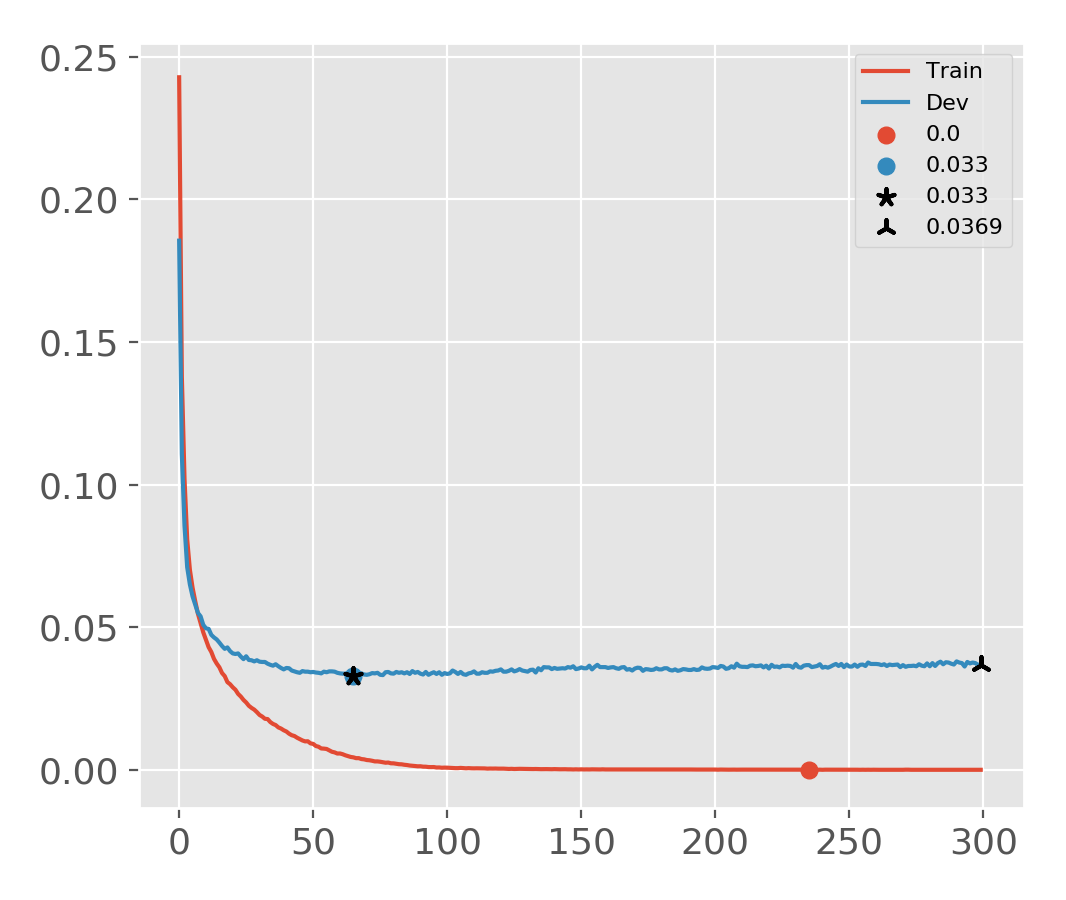}}%
    \subfloat[$l=4095, p=0.05$]{\includegraphics[width=.33\textwidth]{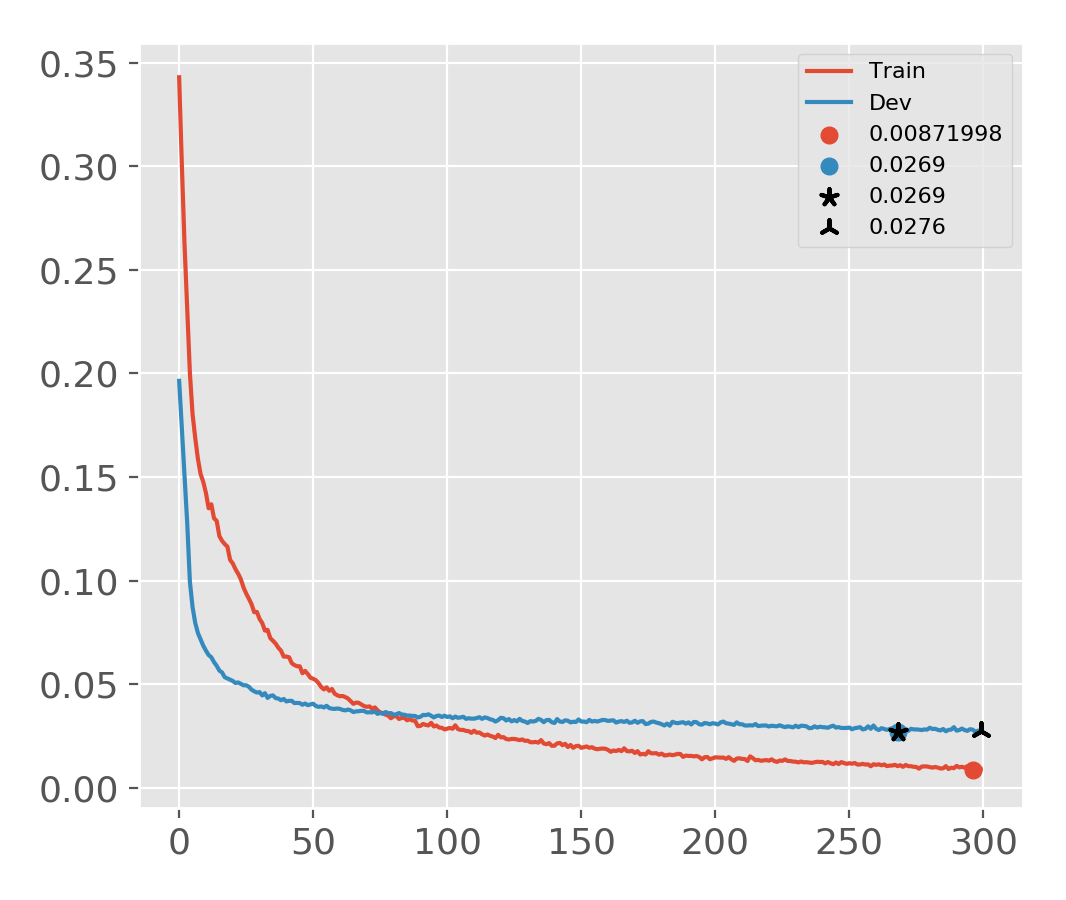}}%
    \subfloat[$l=4095, p=0.1$]{\includegraphics[width=.33\textwidth]{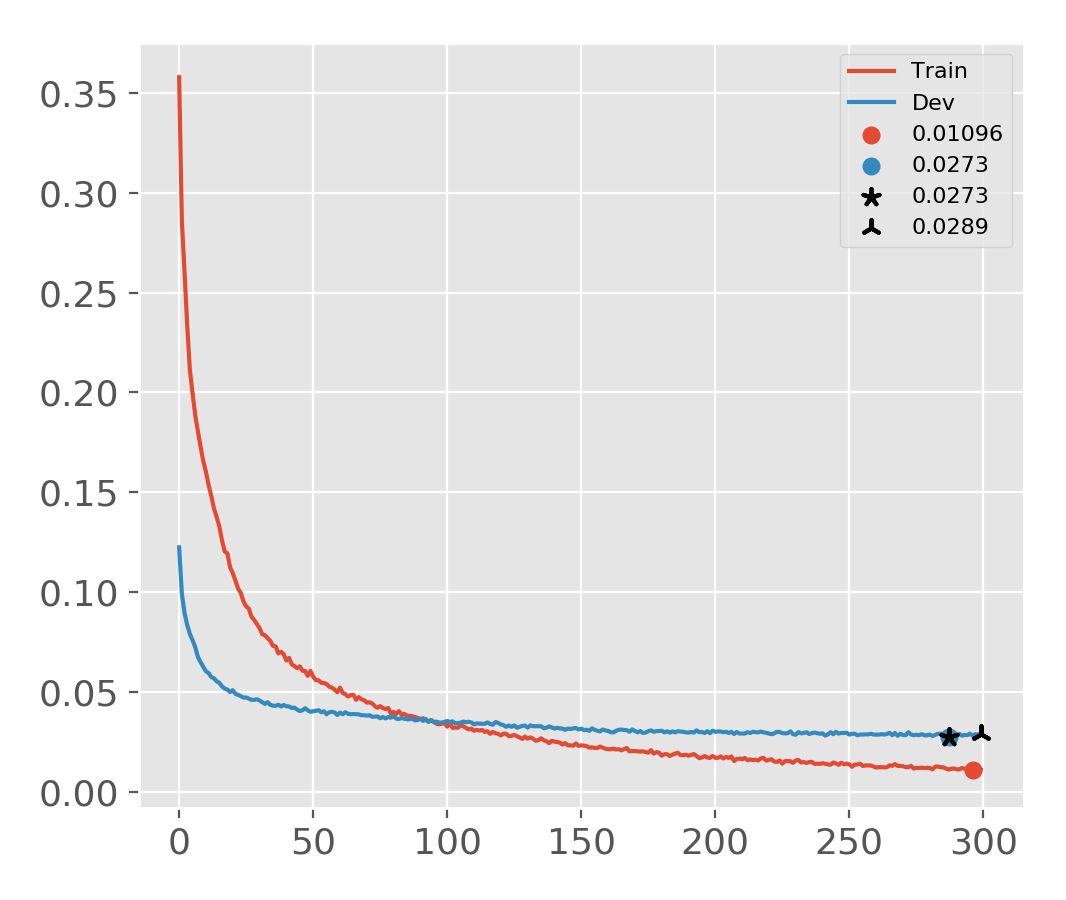}}\\
    \subfloat[$l=8191$, no dropout]{\includegraphics[width=.33\textwidth]{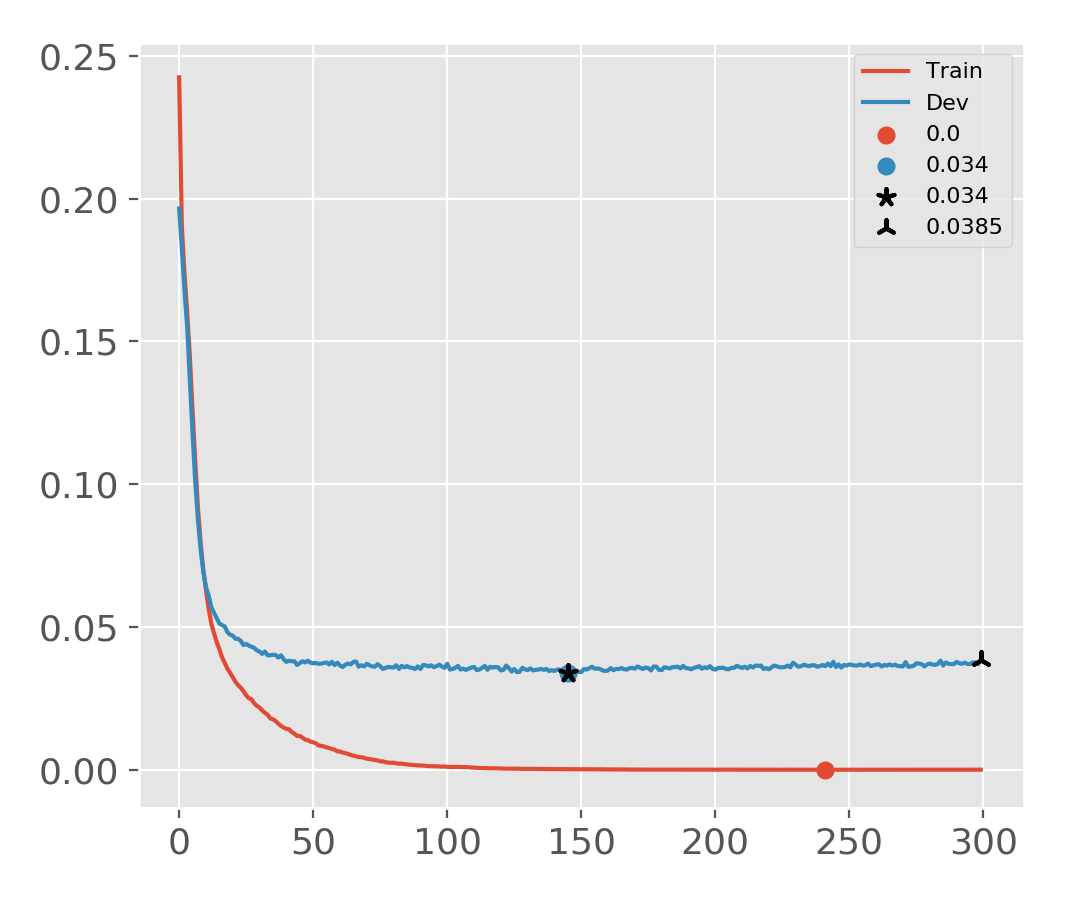}}%
    \subfloat[$l=8191, p=0.05$]{\includegraphics[width=.33\textwidth]{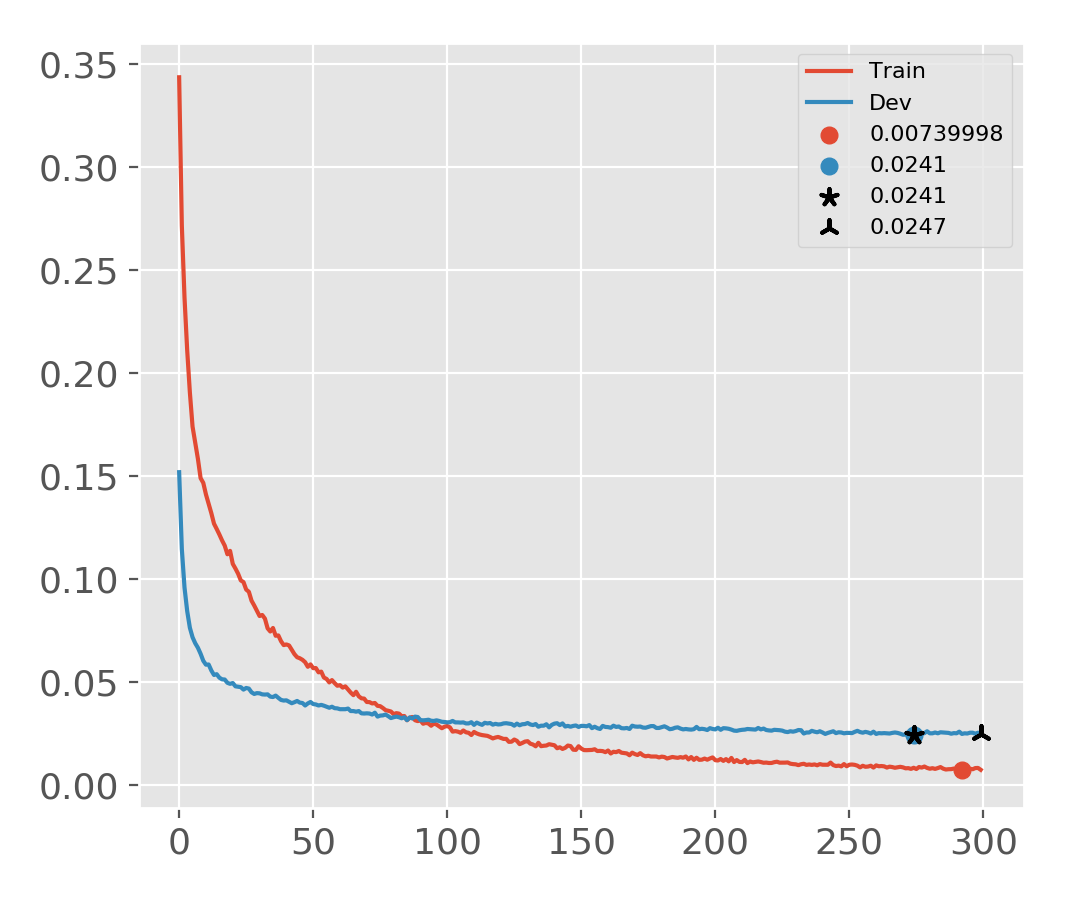}}%
    \subfloat[$l=8191, p=0.1$]{\includegraphics[width=.33\textwidth]{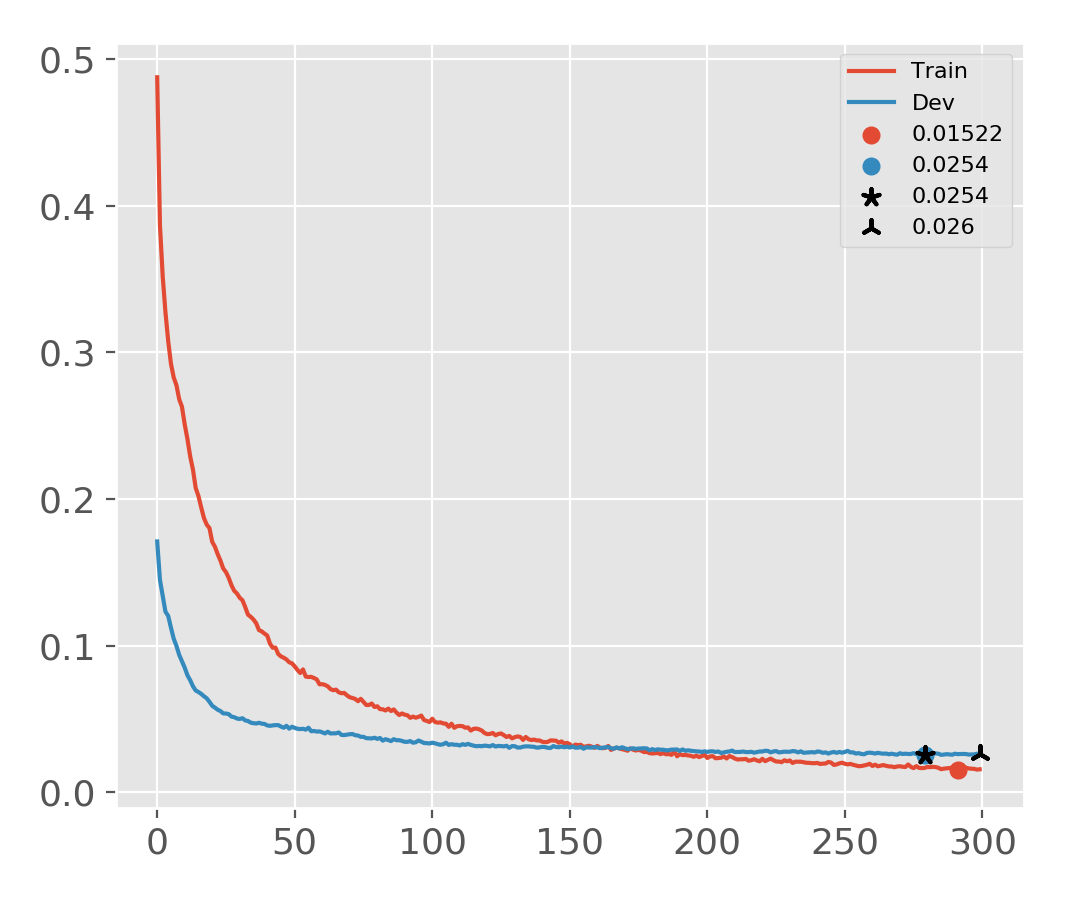}}%
    \caption{Misclassification errors on MNIST for different number of internal nodes $l$ and different dropout rates $p$.
    Red and blue curves show training and validation errors, respectively, where red and blue dots show the minimum
    values for each. Asterisk shows the test set error for the best validation performing model and three-pointed star
    shows the test set error for the very final model.}
    \label{fig:mnist}
\end{figure}

Next, we evaluate our approach on MNIST data which contains 60,000 training and 10,000 test examples of handwritten digit images~\citep{lecun1998mnist}. Each image is $28\times 28$ pixels with single (black \& white) channel (784-dimensional). Output labels are the ten digits, therefore the task is formulated as a 10-class classification problem. We randomly partition the original training set into training and validation sets with a ratio of 5:1. We use the cross-entropy loss with a softmax output at the top. We train for a total of 300 epochs.

Results are presented in Figure~\ref{fig:mnist}. We show misclassification errors for two architectures that have a total number of internal nodes 4095 (depth 12) and 8191 (depth 13) for dropout rates of 0, 0.05 and 0.1.

We see that for both sizes of 4095 and 8191, with no dropout, there is overfitting---the blue curve (validation error) is higher than the red curve (training error), and the gap gets smaller with dropout. In terms of validation performance, a dropout rate of 0.05 beats 0.1, which beats the case with no dropout (for models with 4095 nonleaves, 2.69\% < 2.73\% < 3.3\%; with 8191 nonleaves, 2.41\% < 2.54\% < 3.4\%). The best model ends up being the larger model with a dropout rate of 0.05. In the case of no dropout, not only there is a larger gap between training and validation performances but the validation error also quickly starts curving upwards. In contrast, with dropout, there is both a smaller gap and a consistently improving validation performance---the best model (shown with an asterisk) has lower error and is achieved later. 

\begin{figure}[t]
    \centering
    \subfloat[No dropout]{\includegraphics[width=.49\textwidth]{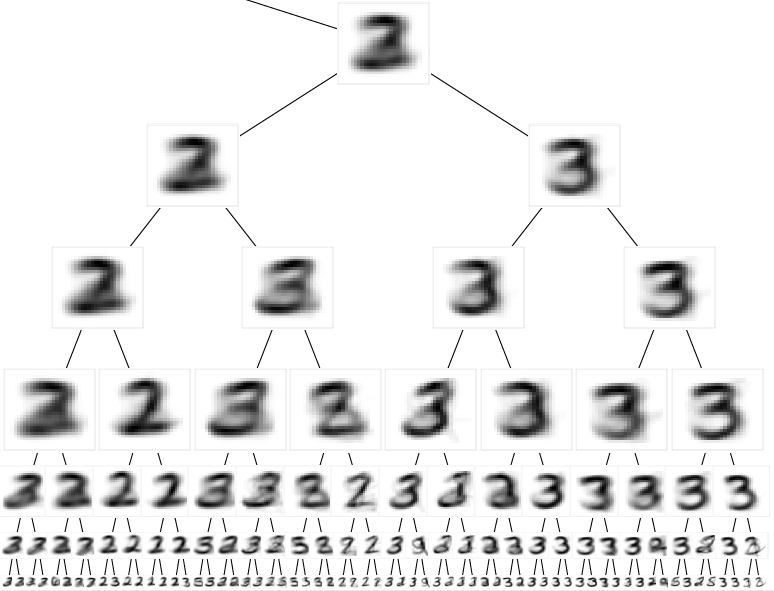}}%
    \subfloat[With dropout]{\includegraphics[width=.49\textwidth]{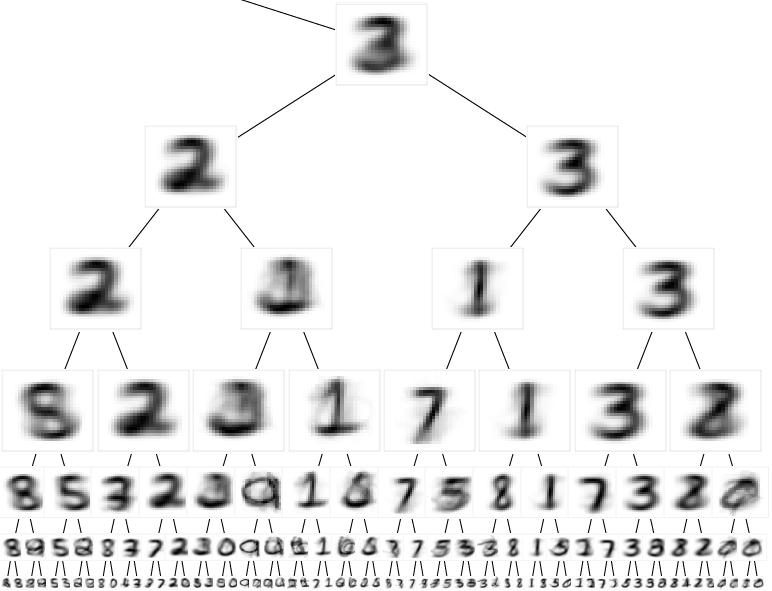}}%
    \caption{A visualization of rightmost subtrees of a 2047- nonleaf hierarchical mixture of experts model.}
    \label{fig:mnisttree}
\end{figure}

\begin{figure}[ht!]
    \centering
    \subfloat[$l=4095$, no dropout]{\includegraphics[width=.33\textwidth]{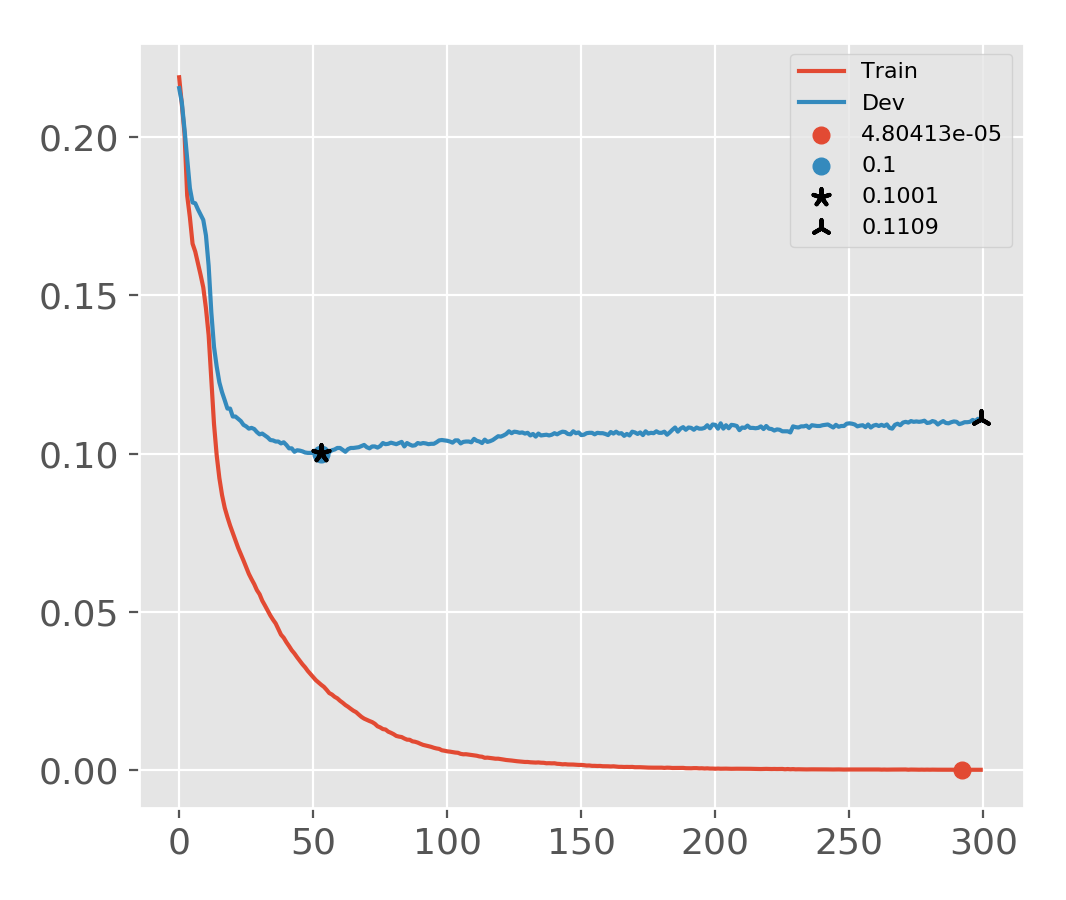}}%
    \subfloat[$l=4095, p=0.05$]{\includegraphics[width=.33\textwidth]{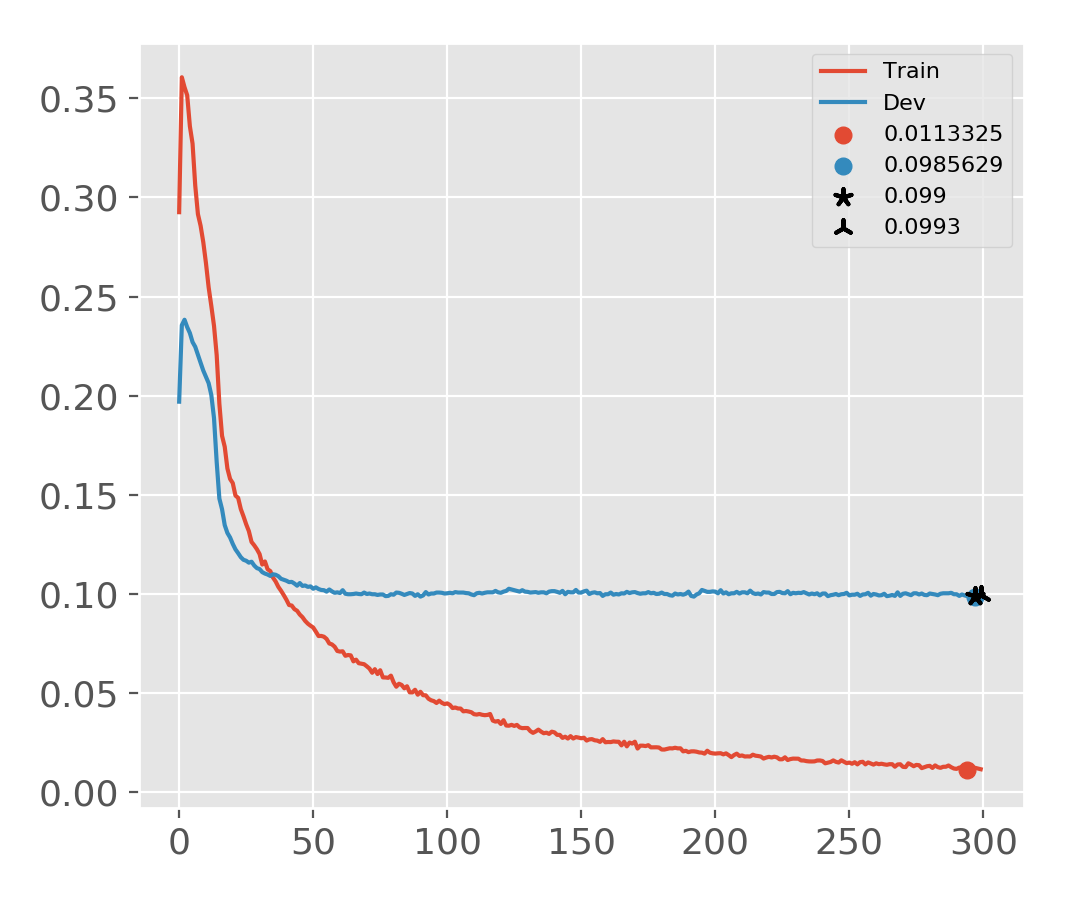}}%
    \subfloat[$l=4095, p=0.1$]{\includegraphics[width=.33\textwidth]{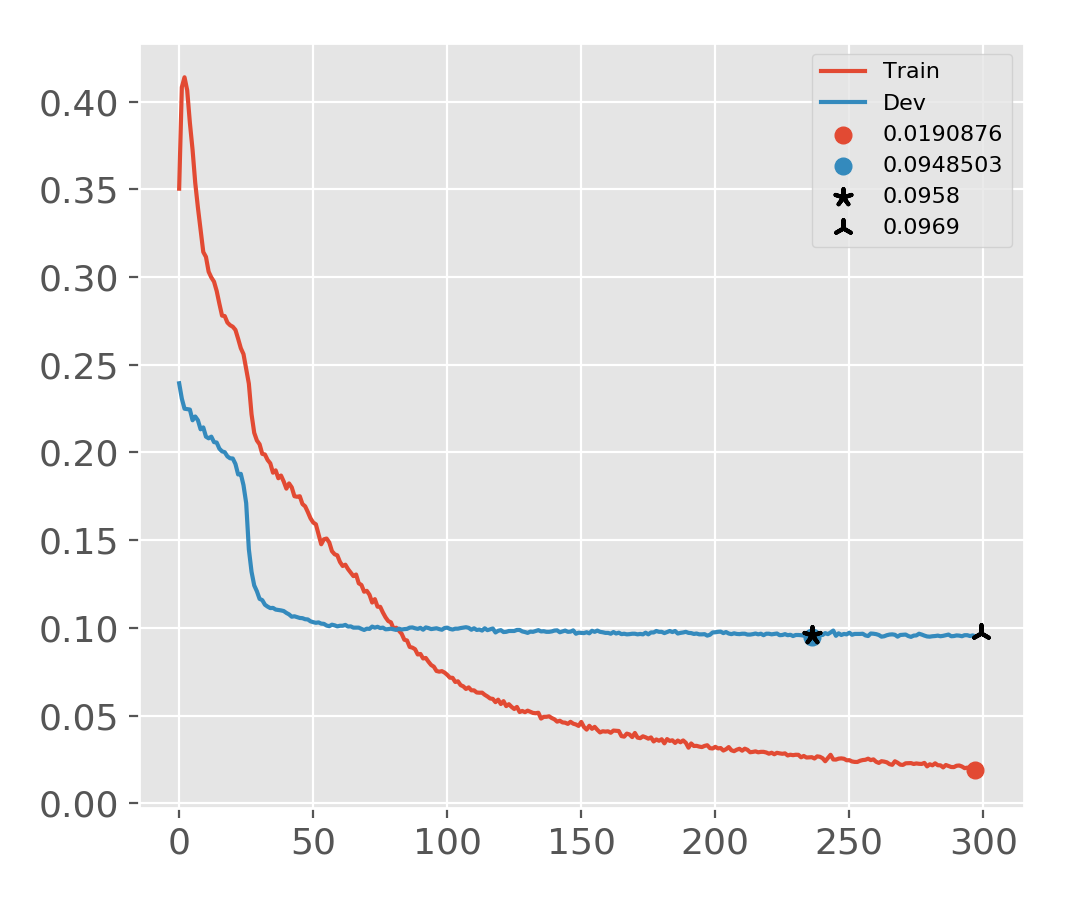}}\\
    \subfloat[$l=8191$, no dropout]{\includegraphics[width=.33\textwidth]{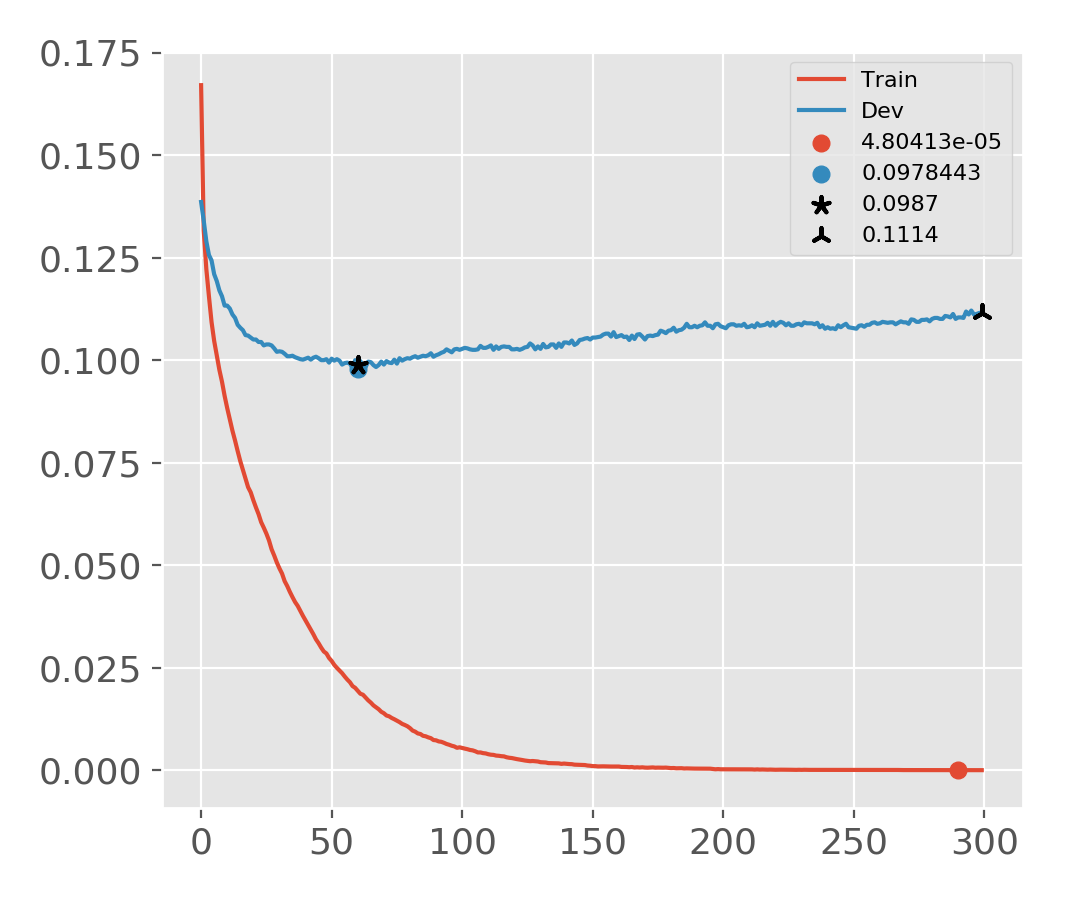}}%
    \subfloat[$l=8191, p=0.05$]{\includegraphics[width=.33\textwidth]{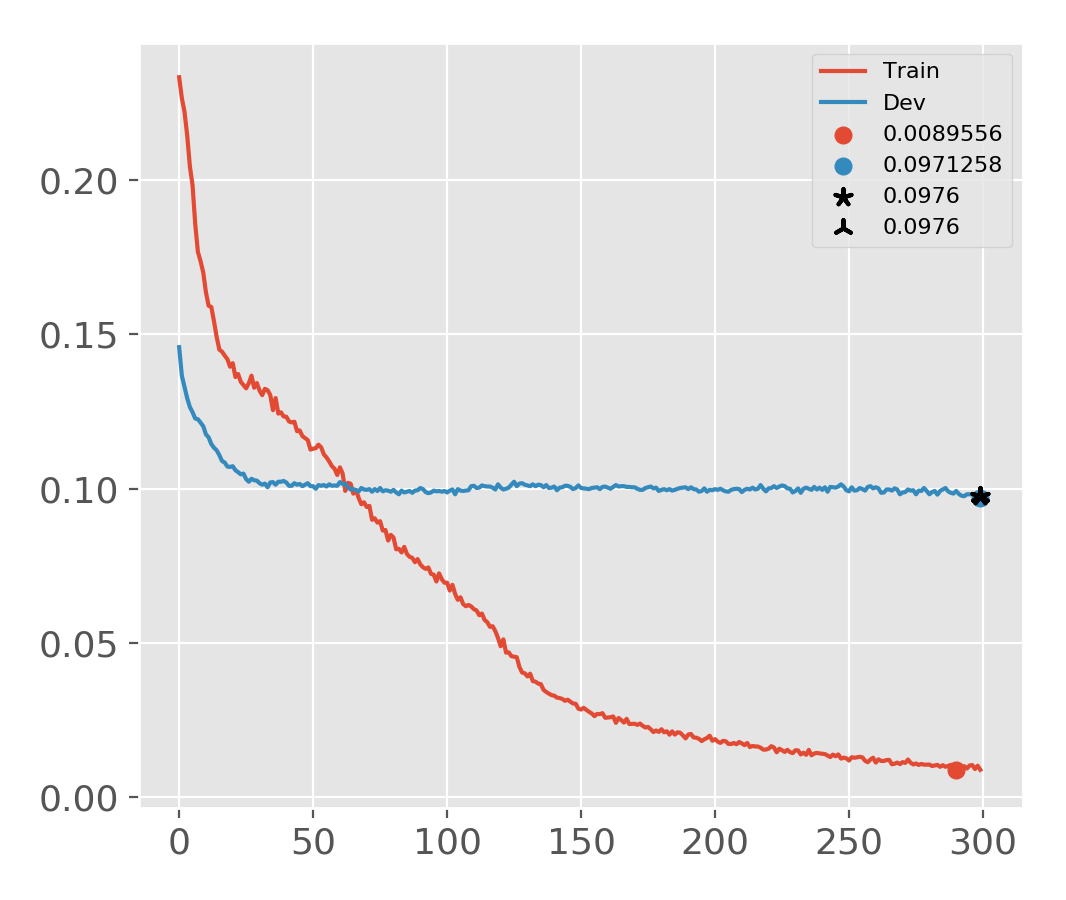}}%
    \subfloat[$l=8191, p=0.1$]{\includegraphics[width=.33\textwidth]{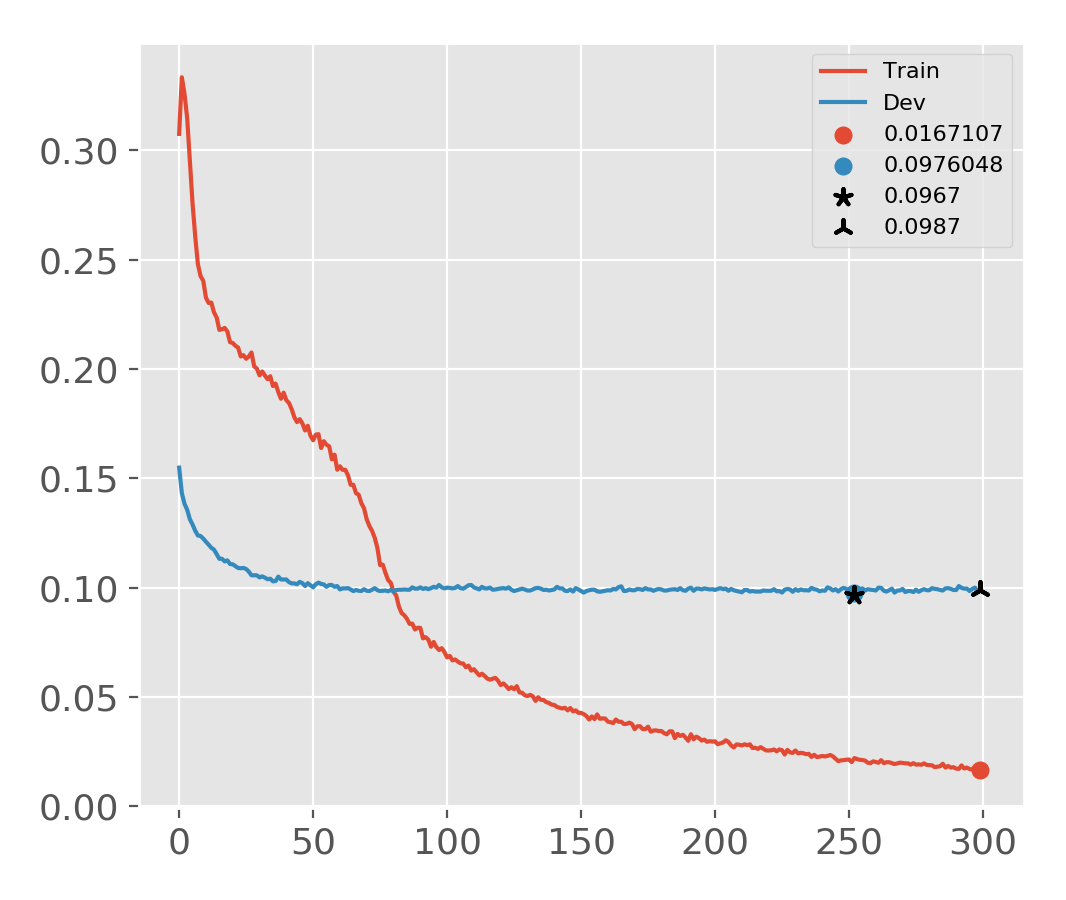}}%
    \caption{Misclassification errors on CIFAR-10 for different number of internal nodes $l$ and different dropout rates $p$. Red and blue curves show training and validation errors, respectively, where red and blue dots show the minimum values for each. Asterisk shows the test set error for the best validation performing model and three-pointed star shows the test set error for the very final model.}
    \label{fig:cifar}
\end{figure}

Additionally, to qualitatively inspect our models, we visualize the learned hierarchy by measuring which input instances activate a given node. A subtree of the resulting visualization is given in Figure~\ref{fig:mnisttree}. We simply take a weighted average over all the input vectors, where weights are determined by the gating activation of that particular node, and treating the resulting that average as the representation learned by that node. We observe that with no dropout, the features tend to look very similar for close siblings and vary very little. In contrast, when we apply dropout, there is more variation in the features. This behavior is intuitive: Whenever we choose to drop, we rely on only a single (right) subtree to implement the prediction function, which forces the right subtree to handle more classes which is reflected by higher variance in the features.

When there is no dropout, the task is distributed over the whole tree; with dropout however with always dropping out the left side branch, we are effectively training an ensemble of trees of different depths and complexities.

\subsection{Results on Image Classification}

We evaluate our models also on the CIFAR-10 dataset which has 60,000 images of $32\times 32$ images with three color channels, each tagged with a class label out of ten classes~\citep{krizhevsky2009learning}. We use the training-test set partitioning with the ratio of 5:1, given by the authors. In addition to the existing training-test split, we randomly partition the training set into training and validation (development) sets with a ratio of 4:1 for early stopping. To represent the images, we use the version 3 of the Inception network~\citep{szegedy2016rethinking}, pretrained on ImageNet~\citep{krizhevsky2012imagenet}. In particular, after running the network on each image, we extract the response from the third pooling layer, which results in a 2,048-dimensional vector for each image instance. Since this task is formulated as a classification problem, we optimize for cross-entropy objective with a softmax nonlinearity at the very top of the model.
We train each model for 300 epochs.

Results are presented in Figure~\ref{fig:cifar}. We show misclassification errors for two architectures that have a total number of 4095 and 8191 nonleaves for dropout rates of 0, 0.05 and 0.1.

When there is no dropout, both architectures seem to get $\sim 10\%$ error, and we see that the validation performance quickly starts to worsen and curve upward, while the training error reaches very close to
zero\footnote{Training error typically starts off worse than development error, which is because the training error is computed in an on-line fashion as the training continues within an epoch. Furthermore, it includes the stochasticity due to dropping out units.}. As we increase the dropout rate, we observe that the gap between training and validation performances shrinks.

For the dropout rate values we have investigated (0, 0.05 and 0.1), we see that increasing the dropout rate consistently improves test performance, which shows the effectiveness of using dropout. For dropout rate of 0.05,
the best validation set performing model is the last model, which shows that overfitting still has not started after 300 epochs and there might be room for further improvement. In contrast, for the case where dropout is not applied, the last model is noticeably worse than the best validation model which is attained at approximately 50th epoch.

\section{Conclusions}

We provide a novel dropout mechanism that can be applied to the hierarchical mixture of experts method and its extensions. In contrast to the dropout on flat architectures with units that are conditionally independent, our method works faithfully to the gating dependencies that exist in the tree hierarchy of the model. 

We show the effectiveness of our approach on a synthetic toy dataset as well as two real-world datasets for the task of digit recognition and image classification. On all data sets, we see that the hierarchical mixture of experts does overfit when there are too many levels and leaves, but that our proposed method works as an effective regularizer with the dropout rate as the hyperparameter that trades off bias and variance. 

We also qualitatively evaluate the impact of dropout on the representations learned by the models visualizing how dropout affects by providing visualizations. Because of the fact that we asymmetrically drop only the left subtree, our dropout method effectively samples from an ensemble of tree-structured models of different complexities. This approach introduces regularization by acting as an interpolation of models having varying complexities.

\end{document}